\documentclass[conference]{IEEEtran}
\usepackage{cite}
\usepackage{amsmath,amssymb,amsfonts}
\usepackage{algorithmic}
\usepackage{graphicx}
\usepackage{wrapfig}
\usepackage{textcomp}
\usepackage{xcolor}
\usepackage{float}
\usepackage{tikz}
\usepackage{adjustbox}
\usepackage{subfig}
\usepackage{url}

\def\BibTeX{{\rm B\kern-.05em{\sc i\kern-.025em b}\kern-.08em
    T\kern-.1667em\lower.7ex\hbox{E}\kern-.125emX}}
\begin{document}
\bstctlcite{IEEEexample:BSTcontrol}

\title{Improving Sample Efficiency and Multi-Agent Communication in RL-based Train Rescheduling}

\author{Dano Roost$^{\dag}$, Ralph Meier$^{\dag}$, Stephan Huschauer$^{\dag}$, Erik Nygren$^{\star}$, Adrian Egli$^{\star}$,\\ Andreas Weiler$^{\dag,\diamond}$, and Thilo Stadelmann$^{\dag,\diamond}$\\ \\
\parbox{4.5 cm}{\centering $^{\diamond}$ ZHAW Datalab\\ {\footnotesize \tt \{wele, stdm\}@zhaw.ch}} 
\parbox{8 cm}{\centering $^{\dag}$ ZHAW School of Engineering\\ {\footnotesize \tt \{roostda1, meierr18, huschste\}@students.zhaw.ch}} 
\parbox{4.5 cm}{\centering $^{\star}$ Swiss Federal Railways\\ {\footnotesize \tt name.surname@sbb.ch}}
}


\maketitle

\begin{abstract}
We present preliminary results from our sixth placed entry to the \emph{Flatland} international competition for train rescheduling, including two improvements for optimized reinforcement learning (RL) training efficiency, and two hypotheses with respect to the prospect of deep RL for complex real-world control tasks: first, that current state of the art policy gradient methods seem inappropriate in the domain of high-consequence environments; second, that learning explicit communication actions (an emerging machine-to-machine language, so to speak) might offer a remedy. These hypotheses need to be confirmed by future work. If confirmed, they hold promises with respect to optimizing highly efficient logistics ecosystems like the Swiss Federal Railways railway network.
\end{abstract}


\vspace{-5pt}
\section{Introduction}

The Swiss Federal Railways (SBB) railway network is frequented more and more due to increased public and freight transport demand. This higher traffic density makes it more difficult to mitigate the effect of small delays of one train on other trains’ schedules by traffic dispatching, leading to secondary delays. Such dispatching is conducted by changing the speed and/or departure times of trains, or by rerouting trains. While this is done manually today, the steadily increase in traffic density necessitates at least a semi-automatic solution leveraging the power of data science \cite{stadelmann2013applied}. However, the amount of switches, tracks, and trains leads to a combinatorial explosion of rerouting options such that a full optimization of the dispatching problem is infeasible. Therefore, SBB has created \emph{Flatland}, a simulation environment and international data science competition to solicit research into the area of multi-agent reinforcement learning as an alternative \cite{nygren2019flatland}.

\begin{wrapfigure}{R}{4.2cm}
    \vspace{-10pt}
    \includegraphics[scale=0.23]{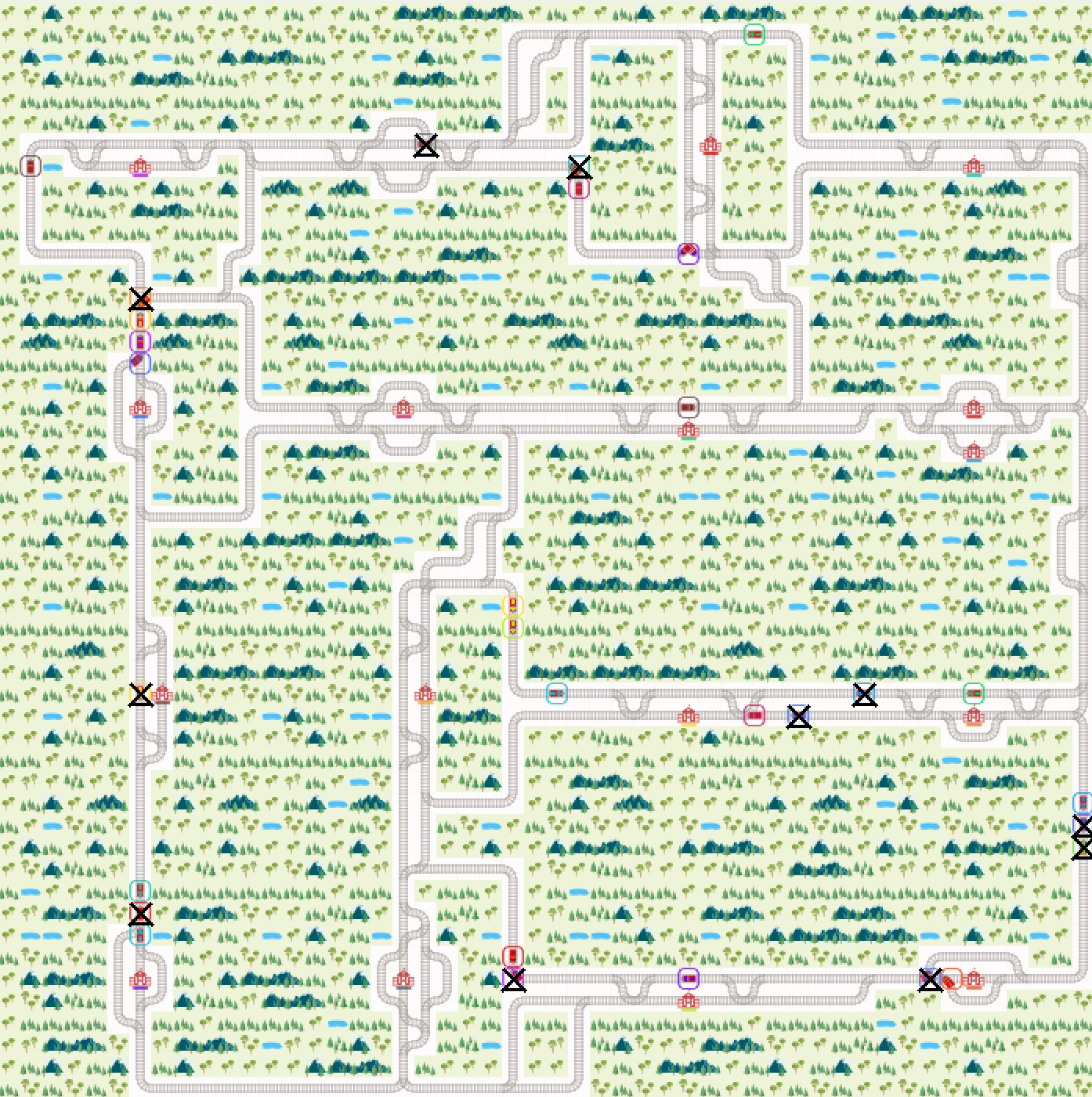}
    \caption{Example environment.}
    \label{fig:flatland_environment} 
\end{wrapfigure}

Goal in the \emph{Flatland} challenge is to successfully guide all trains to their assigned target stations in any instance of a randomized grid-like environment (\textit{cf.} Figure \ref{fig:flatland_environment}), by means of modeling each train as its own RL-trainable agent. This is challenging because a single wrong decision can cause a chain reaction that makes it impossible for many other trains to reach their destinations. A solution to \emph{Flatland} would have implications for any rescheduling problem and related complex transportation tasks as faced by e.g. production and logistics companies. The challenge consists of two rounds: round one focuses on avoiding conflicts with multiple trains (agents) in a number of previously unseen environments. Round two aims at optimizing more realistic traffic including trains with different speed profiles or malfunctions, fewer switchover facilities and more trains in less time.

\setlength{\belowcaptionskip}{-15pt}

In this paper, we present preliminary results from our contribution to the \emph{Flatland} challenge, including two improvements to a more sample efficient training process based on constraining the decision space and applying curriculum learning, respectively, in Section \ref{sec_approach}. Based thereon, we present the analysis of two hypotheses with respect to deep RL for train rescheduling in Section \ref{sec_hypotheses}: \emph{(a)} that despite their general advantages, state of the art policy gradient methods might be inappropriate in environments like \emph{Flatland} due to the stochasticity of the learned policy; and \emph{(b)} that an explicit communication protocol between agents for negotiating precedence on bottlenecks could emerge through RL that offers a remedy for the remaining challenges. Both hypotheses need to be confirmed by more systematic experimental exploration in future work, but hold promises for other complex real-world control tasks involving multiple agents \cite{stadelmann2018deep}.

\section{A Sample-Efficient Approach}
\label{sec_approach}

We use the model-free A3C \emph{reinforcement learning algorithm} \cite{mnih2016asynchronous} for its well-known robustness and learning speed (due to multiple simultaneous training instances). We design our \emph{observation space} as follows: to focus the attention of an agent on the information relevant for its decision making, we use a binary tree that represents all railway ``sections'' ahead of the agent up to a specified depth (set to $3$ for all experiments below). A section is thereby defined as the part of the railway network between two usable switches from the agent’s perspective. Each section represents a node of this binary observation tree and has attributes such as length, number of trains on this section, distance to the agent's target, etc. These values are normalized into the range $0$--$1$ and the tree is flattened into a fixed size vector, which gets concatenated with information about the agent itself like current speed, orientation or malfunction (similar to \cite{bacchiani2019microscopic}). 

To map the perception history of an agent onto the next action, we use a \emph{neural network architecture} with fully connected and recurrent layers. We observe stable training performance with three fully connected layers of sizes $128$--$64$--$64$, with one LSTM layer of size $64$ between the last fully connected ones. An ablation study shows that the version with LSTM yields approximately $11$\% better arrival rates than a version without.

To improve the \emph{training process} in terms of convergence speed beyond the innate parallelization capabilities of A3C, we reduce the number of decisions an agent has to take by defining that as long as the agent is not facing a switch, the default action ``straight ahead'' is taken. This way, the agent only sees perceptions for training where a decision is necessary. In a test environment of size $100$x$100$ with $14$ individual agents, this improves the arrival rate from $44.5$\% to $82.9$\%. To further improve training speed, we use a form of \emph{curriculum learning} that increases the size of the training environments and the number of agents during training. In an experiment, we could confirm that the initially uninformed policy was not able to learn from scratch on large environments but needed to master basic abilities like path finding and basic collision avoidance first, which could be learned upfront on smaller environments.

An early version of our approach without above training process optimizations achieved a reasonable $18^{th}$ rank (out of $37$ participants) in round one of the \emph{Flatland} competition with a submission score (fraction of arrived agents) of $0.489$. Interesting are the \emph{results} for the more complex round two and our full system: the experiments show a strong performance increase, primarily caused by said reduction of the decision space. Due to the larger environments, denser traffic and very sparse grid layouts in the second round, the solution could still only achieve a final score of 29.1\%, surfacing as rank $6$ of $32$ on the leaderboard. We identify the lack of planning capabilities as the main source of error: while collision avoidance in trivial cases is handled well, situations with many agents involved often lead to trains getting stuck.

\section{Hypotheses for Future Work}
\label{sec_hypotheses}

Besides these promising results, our experiments give rise to the following two hypotheses with potentially more far-fetching consequences, hence awaiting further study:  

\paragraph*{Hypothesis 1---general inappropriateness of policy gradient methods in high-consequence environments} Policy gradient-based algorithms like A3C learn a probability distribution over all available actions given a perception. Let environments like \emph{Flatland}, where taking one bad action quickly leads to a chain reaction of unresolvable situations, be called \emph{high-consequence environments}. The difficulty of combining a high-consequence environment with a stochastic policy can be illustrated with an example: if $10$ agents in an environment choose their best action with a probability of $90$\%, leaving $10$\% likelihood for non-beneficial and potentially catastrophic actions, there is already a $65.1$\% chance that one of the agents takes an action that might create a chain reaction of problems. Just converting this probability distribution into a deterministic policy by taking the $\arg\max$ over the distribution does not solve the problem due to some situations in which the agent is not sure what to do and therefore assigns similar probabilities to different actions. Policy gradient methods then rely on their stochasticity to try all available actions during training. While in most popular RL use-cases such as Atari games, it will suffice to select a good action and not necessarily the best, this is different in many real-world tasks and should therefore be addressed. An option would be to experiment with purely value-based RL algorithms to observe if such methods can overcome the described problem of policy gradient methods.

\setlength{\belowcaptionskip}{-20pt}
\begin{figure}[t]
    \centering
    \includegraphics[scale=0.4]{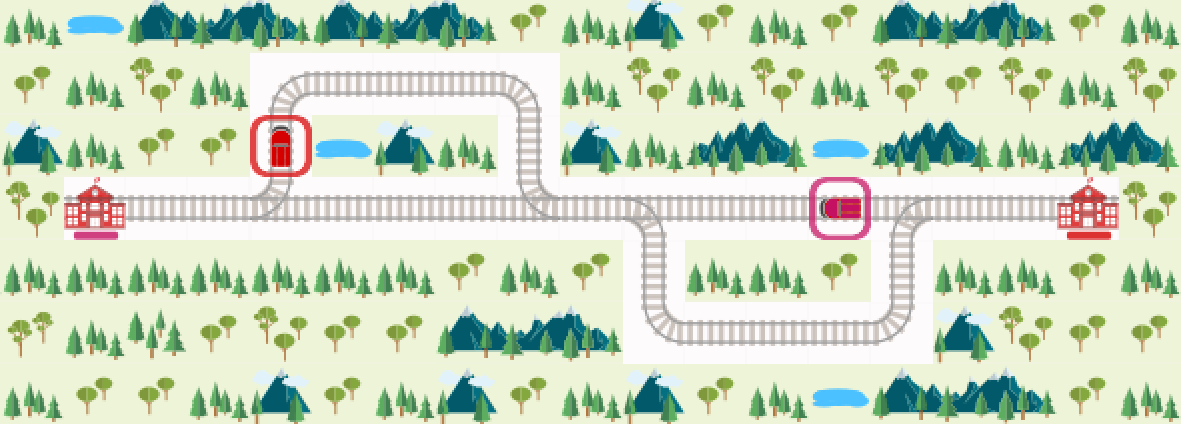}
    \caption{Environment for communication experiment.}
    \label{fig:communication_experiment} 
\end{figure} 

\paragraph*{Hypothesis 2---learning to communicate pays off} Consider Figure \ref{fig:communication_experiment}, where two trains starting at opposite ends of the environment need to switch their positions through a single-track section. Human operators could solve the task by \emph{communicating} with each other, negotiating who would take the detour in order to let the other pass straight. In an experiment, we gave our system the opportunity to learn such behavior in principle by adding five new ``communication actions'' (undetermined in their meaning), a sixth \texttt{EOT} (``end of transmission'') action and a shared ``communication buffer'' within the observation space. On taking a communication action, the two agents start a communication loop that allows the two agents to alternately read from the buffer, calculate an action and write that action back into the buffer, until both agents output \texttt{EOT}. Then, both agents can select a regular action to proceed in the environment. If both agents take the same action and collide, we give a reward of $-1$. If they make it around each other and reach their targets, they receive a reward of $+1$. The agents have no way to know which agent they are (hence, cannot learn to always go one specific way). After $100'000$ episodes of training, the agents are able to solve the task in $95$\% of the cases compared to $47$\% without communication. Interestingly, we observe that to a large degree the communication is not repetitive between episodes, but highly variable (most episodes require between one and $4$ communication rounds). Future work should therefore shed more light on the properties of the language emerging between the agents and its use in more realistic scenarios.


\bibliographystyle{IEEEtran}

\vspace{-6pt}
\bibliography{main}

\end{document}